# Towards Non-Toxic Landscapes: Automatic Toxic Comment Detection Using DNN

**Ashwin Geet D'Sa, Irina Illina, Dominique Fohr**

Université de Lorraine, CNRS, Inria, LORIA

F-54000 Nancy, France

**Abstract**

The spectacular expansion of the Internet has led to the development of a new research problem in the field of natural language processing: automatic toxic comment detection, since many countries prohibit hate speech in public media. There is no clear and formal definition of hate, offensive, toxic and abusive speeches. In this article, we put all these terms under the umbrella of "toxic speech". The contribution of this paper is the design of binary classification and regression-based approaches aiming to predict whether a comment is toxic or not. We compare different unsupervised word representations and different DNN based classifiers. Moreover, we study the robustness of the proposed approaches to adversarial attacks by adding one (healthy or toxic) word. We evaluate the proposed methodology on the English Wikipedia Detox corpus. Our experiments show that using BERT fine-tuning outperforms feature-based BERT, Mikolov's and fastText representations with different DNN classifiers.

## 1. Introduction

The past few years have seen a tremendous rise in the usage of Internet and social networks. Unfortunately, the dark side of this growth is an increase in toxic speech. Toxic speech is a type of offensive communication mechanism. Toxic speech can target different societal characteristics such as gender, religion, race, disability, etc. (Delgado and Stefancic, 2014) and reflects a certain "state of society". There is no uniform definition of toxic speech in the scientific literature and there is no clear distinction between *hate, offensive, toxic and abusive speech* (Gröndahl *et al.*, 2018; Waseem *et al.*, 2017; Davidson *et al.*, 2017). We refer to these collectively with the generic term of *toxic* speech.

Manually monitoring and moderating the Internet and social media content to identify and remove toxic speech is extremely expensive. This article aims at designing methods for automatic toxic speech detection on the Internet. Despite the studies already published on this subject, the results show that the task remains very difficult (Nobata *et al.*, 2016; Saleem *et al.*, 2017). In this paper, we use semantic content analysis methodologies from *Natural Language Processing* (NLP) and methodologies based on *Deep Neural Networks* (DNN).

Very recently, DNNs have become the state-of-the-art method for toxic speech detection. Badjatiya *et al.* (2017) investigated the application of DNNs for hate speech detection and compared it with various classical features like character n-grams, Term Frequency-Inverse Document Frequency (TF-IDF) values, Bag of Word Vectors (BoWV), and Global Vectors for Word Representation (GloVe) (Pennington *et al.*, 2014). They found DNN methods to significantly outperform the existing shallow methods. Zhang *et al.* (2018) combined Convolutional neural network (CNN) and Recurrent neural network (RNN) by giving the output of CNN to RNN with Gated Recurrent Unit (GRU). Van Aken e*t al*. (2018) proposed a combination of shallow models and DNN methods that outperforms all the individual models. Several evaluations of a range of NLP features was performed by Nobata *et al*. (2016). Stammbach *et al*. (2018) reported different pre-processing techniques and their impact on the final classification. Wulczyn *et al*. (2017) went beyond the simple classification task and developed a method that combines crowdsourcing and machine learning to analyse personal attacks.

Currently, one of the most powerful semantic context representations are those obtained from BERT (*Bidirectional Encoder Representations from Transformers*) (Devlin *et al.*, 2019; Young *et al.*, 2018). Compared to Mikolov's embedding (Mikolov *et al.*, 2013), BERT model takes into account large left and right semantic contexts of words and can generate different semantic representations for the same word based on its context. Furthermore, pre-trained BERT model can be *fine-tuned* to a specific NLP task (Peters *et al.*, 2019). The BERT model has resulted into new state-of-the-art for several NLP tasks.

In this article, we investigate several approaches based on different state-of-the-art DNN models and word representations for the task of automatic toxic comment detection. Among the classifiers, we used top performing DNNs in the field of NLP: CNN and RNN. CNN allows the extraction of local features in text, e.g. pertinent sequences of words. RNN is able to extract long-term dependencies that are definitely useful for toxic comment detection (Del Vigna *et al.*, 2017). To take into account the semantic context of the document, we propose to use different representations: Mikolov's, fastText and BERT embeddings. We compare these against transformers based BERT fine-tuning. The designed systems are evaluated on publicly available corpus of toxic comments from Wikipedia. The work of Bodapati *et al*. (2019) compares CNN based and fastText classifiers with various character and word based input representations to BERT fine-tuning. As compared to Bodapati *et al*. (2019), we go beyond binary classification and propose a regression-based method. Furthermore, we analyse the robustness of these approaches with adversarial attacks by adding a toxic or healthy word to the comment. Additionally, we have compared CNN based architecture against RNN based Bi-LSTM and Bi-GRU classifiers. It should be noted that our results of binary classification are not directly comparable to Bodapati *et al*. (2019) due to differences in training and pre-processing setup.

The rest of the paper is organized as follows: Section 2 describes the approaches. The experiment protocol and the data are described in section 3. The classification results are discussed in section 4.

## 2. Proposed methodology

Figure 1 presents a schema of our proposed methodology along with the different word representations. We first describe the different word representations and then discuss the DNN classifiers that we evaluate. In all our approaches, the DNN outputs represent the toxicity of a comment.

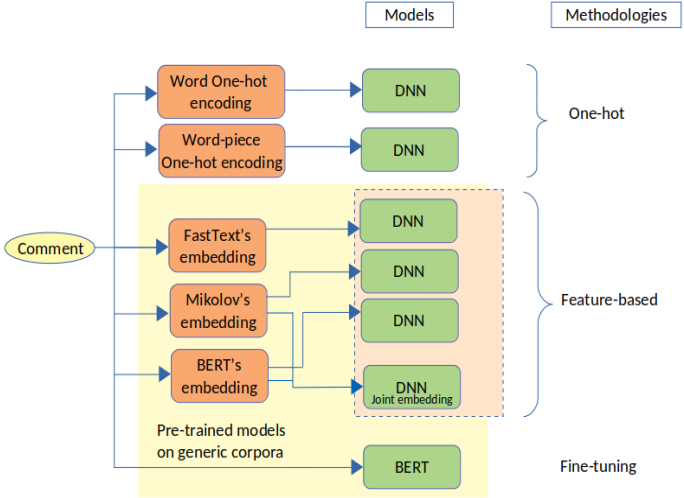

Figure 1: Proposed system architecture for toxic comment detection.

### 2.1 Comment representations

#### 2.1.1 Baseline approach: one-hot representation

Our baseline is the classical one-hot input representation, wherein each input word is represented by a one-hot vector. Only the N most frequent words of training corpus are selected. The other words are represented as UNK. One-hot vectors are used as input to DNN classifier. The DNN will classify these sequences of one-hot vectors as toxic or non-toxic. The first hidden layer of the DNN computes the word embeddings. The weights of this embedding layer are trained together with the weights of the other layers of the network. The particularity here is that we do not exploit any pre-trained word embeddings and the entire training is performed using only the task specific corpus.

#### 2.1.2 Feature-based approaches

Embedding models entail vector-based word representations which are usually pre-trained on large datasets. In this work, pre-trained word representations are used as features in task-specific DNN architectures. The DNN network classifies these sequences of word embeddings as toxic or non-toxic. We study and compare three state-of-the-art unsupervised word embedding models:
- **Mikolov's word embedding**, which represent each word by taking into account a relatively small window of left and right context words (Mikolov *et al*., 2013).
- **fastText subword embedding.** It is an extension of Mikolov's embedding, which takes into account subword information and allows us to include rare and out-of-vocabulary words (Bojanowski *et al*., 2016; Mikolov *et al*., 2018).
- **BERT WordPiece model (Devlin *et al*., 2019).** This model takes into account long left and right contexts of words. Thanks to this model, for each comment, embedding of each word-piece can be computed and used as input for DNN classifier. In the case of BERT model, the same word-piece can have different embeddings depending on the context.

It is important to note that these representations are pre-trained on corpora not specific to our task of toxic comment detection. Hence, will not be efficient to model the specificity of toxic speech (slang, affronts, abuse, etc.).

#### 2.1.3 BERT fine-tuning approach

The principle of fine-tuning consists in starting from a pre-trained model and updating the model parameters on the task specific corpus. As our task of hate speech detection is an NLP task where context plays a critical role, the architecture of BERT will be very appropriate. We take the same BERT pre-trained model as in Section 2.1.2 and we fine-tune this model using our training data. For fine-tuning, the hyper-parameters: batch size, learning rate, and the number of the training epochs are varied.

### 2.2 DNN classifiers

The task of toxic comment detection can be viewed from two perspectives:
- **A binary classification task:** The neural network is directly trained to decide if a comment is toxic or non-toxic.
- **A regression task:** For each comment we compute a score between 0 and 1 as a normalized average of labels from different annotators. The neural network is trained for predicting these scores (regression task). A threshold on the predicted score can be used to decide if the comment is toxic or not. The threshold is adjusted on the development set to maximize the F1-score.

We investigate three state-of-the-art DNN architectures for our tasks:
- CNN to identify local patterns in the comments;
- bi-directional Long Short-Term Memory (bi-LSTM) to capture long range dependencies in the comments;
- bi-directional Gated Recurrent Unit (bi-GRU), to capture long range dependencies with lesser model parameters;

## 3. Experimental setup

### 3.1 Data description

#### 3.1.1 Wikipedia Detox corpus

We used the data collected in the framework of *Wikipedia Detox* project (Wulczyn *et al*., 2017), including user's talks. In our work we exploited only the **toxicity** part of the corpus. This part contains 160k comments from English Wikipedia talk pages, each labelled by approximately 10 annotators via crowd-sourcing, on a spectrum of how toxic/healthy the comment is with regard to the conversation.

The following toxicity rates are used by annotators: *very toxic, toxic, neither, healthy, very healthy*. According to this label definition, toxic speech corresponds to *very toxic* and *toxic* labels.

For many comments in the Wikipedia Detox corpus, there is a disagreement between annotators. Sometimes, it is difficult to define a dominant label for a comment. To perform the **binary classification** (toxic or not toxic), for each comment, we decided to use the following majority vote labelling:

*if* [(# of very toxic and toxic annotations) >
(# of healthy and very healthy annotations)]
*and* [(# of very toxic and toxic annotations) > 2]
    comment is *toxic*
*otherwise* comment is *non-toxic*

Some examples of the toxic comments are: "*You are a big fat idiot, stop spamming my userspace*", "*What the fuck is your problem?*", "*God damn it fuckers, i am using the god damn sand box*".

### 3.1.2 Train, development and test corpus

We used the train/development/test partition provided with the Detox corpus (respectively 96k, 32k, 32k). Training data is used to train our classifiers and to fine-tune the BERT model. Development corpus is used to tune the hyper-parameters. Test corpus is used to evaluate the performance of the system. We compared the classifier predictions in terms of **F1-score.**

### 3.2 Data pre-processing

For many NLP tasks, training data pre-processing has an important impact on the performance of the system. Moreover, DNN approaches are data-driven. These two factors give a very high importance to the pre-processing.

| Detox corpus | Training | Development | Test |
|---|---|---|---|
| # comments | 88.9K | 32.1K | 31.8K |
| # toxic comments | 16.0K | 5.6K | 5.5K |
| # non-toxic comments | 72.9k | 26.5K | 26.3K |
| Corpus size (word count) | 4.3M | 1.9M | 1.9M |
| # unique words | 106K | 64K | 64K |

Table 1: Statistics on *Wikipedia Detox* data after pre-processing. 'K' denotes thousand, 'M' denotes million.

We decided to set the maximum length of a comment to 200 words for reducing the computation time and for avoiding the out-of-memory problems for BERT (because it is a very large model). For this, we keep the first 200 words of each comment of the training, development and test sets. We removed the toxic comments with more than 200 words per comment from the training set because it is possible that the toxic part of the comment is located after the 200$^{th}$ word. We performed this removal only for training. This pre-processing removed about 5% of toxic comments from the training set. Table 1 shows that toxic comments represent only about 17% of all comments. So, our corpus has an unbalanced class distribution.

We converted all words to lowercase and used uncased BERT, fastText and Mikolov's pre-trained models. We removed the punctuations for the Mikolov's, fastText and one-hot approach. We kept the punctuation for the BERT model.

### 3.3 Embedding models

As Detox corpus is limited in size, we used pre-trained models:
**Mikolov's word embedding:** provided by *Google*[1] and pre-trained on a wide corpus of 100G words from Google news corpus. Embedding dimension is 300 for 3M words.
**fastText subword embedding:** provided by *Facebook*[2] and pre-trained on Wikipedia 2017, UMBC webbase and statmt.org news datasets with total 16B tokens. Embedding dimension is 300, the vocabulary is 1M words.
**BERT-base WordPiece model:** English (uncased) model provided by *Google*, pre-trained on *BookCorpus* and *Wikipedia*, with 12 transformer layers and 12 self-attention heads. The embedding size is 768, the number of WordPieces is 30k (including the punctuations). The total number of parameters is 110 million.

WordPiece BERT model and fastText models succeed to represent all words in our corpus. Mikolov's embedding is a word based model. Some words from our corpus are not included in its vocabulary i.e, *Out-Of-Vocabulary* (OOV) words. Our training set has 86.5k occurrences of OOVs (2%), development set has 45.8k OOV occurrences (2.4%) and the test set has 45.3k (2.4%). To obtain an embedding for these OOV, we compute an average of the embeddings of all the words in the vocabulary.

### 3.4 DNN model configurations

The evaluated configurations are presented in the following: for one-hot approach we keep the 75K or 100K most frequent words. For CNN based model we explored one or two convolutional layers (filter size between 3 and 5), followed by two dense layers (with 64-256, 16-64 dense units), with or without dropout. For bi-LSTM and bi-GRU, we explored one or two layers (with 50, 128 units), followed by one or two dense layers (with 64-256, 16-64 dense units), with or without dropout. We use L2 regularization and *adam* optimizer. For fine-tuning BERT we used maximum sequence length of 256, batch size of 32, learning rate of $2 \cdot 10^{-5}$ and 2 epochs.

## 4. Results and Discussion

### 4.1 Binary classification

Table 2, part A, shows the results for baseline methods for one-hot approach: using words or using the same word-pieces as in BERT. Part B focuses on pre-trained embeddings for feature-based approaches. Moreover, we concatenate Mikolov's and BERT embeddings together and use it as input features to DNN (indicated as '*Mikolov's+BERT word embedding*' in Table 2). In this model, words split into word-pieces by BERT tokenizer are averaged and concatenated with corresponding Mikolov's word embedding. The embeddings obtained by averaging the word-piece tokens are indicated as '*BERT word embedding*'. For '*Mikolov's+BERT fine-tun. word emb.*' configuration we concatenate Mikolov's and BERT fine-tuned embeddings. The results of Part C are obtained by BERT fine-tuning. For the two parts (A, B), we have experimented with three different classifiers: CNN, bi-LSTM and bi-GRU.

As shown in the table, our proposed methods in part B and C show better performance than the baseline methods in part A. Among the classifiers, bi-LSTM and bi-GRU performs slightly better than the CNN. Mikolov's embedding of part B performs worse than one-hot approach. This can be due to the presence of OOV words: the one-hot approach models N most frequent words of training corpus, while Mikolov's embeddings is trained on non-toxic corpus and it is possible that some important toxic words (slang) of our corpus are missing in the Mikolov's pre-trained model. BERT with words (*BERT word embedding*) slightly underperforms compared to BERT word-piece embeddings. This can be due to some loss of information while averaging the embeddings. BERT embedding performed better than one-hot approach. Joint embedding (*Mikolov's+BERT*) give slightly better performance than BERT embedding alone. The best method is BERT fine-tuning which achieves 78.2% F1-score. Joint embedding *Mikolov's+BERT fine-tuned word embedding* achieves the performance close to BERT fine-tuning. Table 2 exhibits that BERT is effective for both the fine-tuning and feature-based approaches. It is worth noting that the evaluated models have a different numbers of learned parameters: DNN based classifier models have about 1M parameters, whereas BERT fine-tuned model has 110M parameters. BERT embedding is a good trade-off between performance and number of model parameters.

---

[1] https://github.com/mmihaltz/word2vecGoogleNews-vectors

[2] https://fasttext.cc/docs/en/english-vectors.html

|  | CNN | bi-LSTM | bi-GRU |
|---|---|---|---|
| A. *One-hot approaches* | | | |
| Word-based | 72.9 | 74.2 | 73.9 |
| Word-piece based | 73.1 | 74.1 | 74.4 |
| B. *Feature-based approaches* | | | |
| Mikolov's embedding | 70.6 | 72.7 | 72.0 |
| fastText embedding | 73.3 | 74.1 | 74.8 |
| BERT embedding | 75.0 | 75.6 | 75.7 |
| BERT word embedding | 74.2 | 75.4 | 75.5 |
| Mikolov's+BERT word emb. | 75.9 | 76.1 | 76.3 |
| Mikolov's+BERT fine-tun. word emb. | 78.0 | 78.0 | 78.0 |
| C. *BERT fine-tuning* | | | |
| BERT fine-tuning | | | **78.2** |

Table 2: Binary classification F1-score for different classifiers and different input representations.

A preliminary error analysis shows that sometimes non-toxic speech can be misclassified as toxic speech in the presence of words like *bullies*, *anti-semitism*. For example, the comment "*You're a nice guy Irishpunktom. It takes guts to speak against bullies.*" is misclassified as toxic. Likewise, toxic speech is misclassified as non-toxic speech due to sarcasm, irony, rhetoric question, etc. For example, "*Thats fine. Thank your extreme rudeness. That front page looks so unwelcoming.*" is misclassified as non-toxic.

### 4.2 Classification using regression model

These experiments compare the performances based on the regression model. A threshold is applied to the regression score to decide if the comment is toxic or not. We use bi-LSTM classifier as it gives the best performance according to Table 2.

We observe that BERT model is more powerful than other models. As for binary classification, BERT fine-tuning gives the best results. Mikolov's+BERT word embedding shows the results close to BERT fine-tuning. We obtained the following results in terms of RMSE *(Root Mean Square Error)* and MAE *(Mean Absolute Error)*:

| | | |
|---|---|---|
| Word-based one-hot | 0.065 | and 0.050; |
| Word-piece based one-hot | 0.065 | and 0.050; |
| Mikolov's | 0.066 | and 0.049; |
| fastText | 0.062 | and 0.047; |
| BERT | 0.062 | and 0.047; |
| Mikolov's+BERT word emb. | 0.06 | and 0.047; |
| BERT fine-tuning | 0.06 | and 0.047. |

These measures further confirm our conclusions.

### 4.3 Robustness evaluation

In order to evaluate the robustness of our classification systems, we added a toxic word ('*fuck*') to each comment of the test set and a healthy word ('*love*') to each comment of the test set. Table 4 shows the percentage of correctly classified comments that change from predicted *non-toxic* to *toxic* comments when a toxic word is appended, and from *toxic* to *non-toxic* when a healthy word is appended. In these experiments, we use bi-LSTM (the best DNN according to Table 2) and the threshold of 0.6 with the regression model. We perform the tests only on feature-based models.

| A. *One-hot approaches* | |
|---|---|
| Word-based | 72.9 |
| Word-piece based | 74.1 |
| B. *Feature-based approaches* | |
| Mikolov's embedding | 74.1 |
| fastText embedding | 75.7 |
| BERT embedding | 76.2 |
| Mikolov's+BERT fine-tun. word emb. | 77.7 |
| C. *BERT fine-tuning* | |
| BERT fine-tuning | **78.0** |

Table 3: F1-score for Bi-LSTM classifier and different input representations using a threshold on regression model.

| | Binary classification | | | Regression model | | |
|---|---|---|---|---|---|---|
| | Mikolov | fast Text | BERT | Mikolov | fast Text | BERT |
| *non-toxic to toxic* | 88.0 | 78.1 | 37.5 | 71.9 | 78.0 | **34.1** |
| *toxic to non-toxic* | 6.5 | 4.8 | **4.1** | 10.9 | 10.0 | 7.6 |

Table 4: Percentage of correctly classified comments, a new word is appended. Bi-LSTM and different models.

We observe that all models are susceptible to the word appending attacks, as also observed in (Gröndahl *et al.*, 2018). Classifiers using Mikolov's and fastText embeddings are more sensitive to appending of a single word. Classifier using BERT embedding is more robust.

## 5. Conclusion

In this article, we have investigated several approaches for toxic comment classification using DNNs. We explored feature-based unsupervised comment representations using Mikolov's, fastText and BERT pre-trained models. These representations are used as input for DNN networks. These approaches are compared to the BERT fine-tuning. We designed binary classification and regression-based approaches. On Wikipedia Detox corpus, our analysis has shown that BERT fine-tuning is the most efficient at this task. Moreover, BERT embedding is the most robust to word attacks. Among DNN based classifiers, bi-LSTM performs better than CNN and bi-GRU at classifying toxic speech.

In the future, we would like to study the impact of data bias on toxic speech detection (Wiegand *et al.*, 2019) and to perform depth study of the multi-class classification (Vaswani *et al.*, 2017). A detailed error analysis to evaluate the linguistic phenomena will also be performed. Moreover, models like XLNet pre-trained model (Yang *et al.*, 2019) or ULMFiT pre-trained language model (Howard and Ruder, 2018) can be studied.

24

## 6. Acknowledgements

This work was funded by the M-PHASIS project supported by the French National Research Agency (ANR) and German National Research Agency (DFG) under contract ANR-18-FRAL-0005.